\documentclass{article}

\usepackage{PRIMEarxiv}

\usepackage[utf8]{inputenc} % allow utf-8 input
\usepackage[T1]{fontenc}    % use 8-bit T1 fonts
\usepackage{hyperref}       % hyperlinks
\usepackage{url}            % simple URL typesetting
\usepackage{booktabs}       % professional-quality tables
\usepackage{amsfonts}       % blackboard math symbols
\usepackage{nicefrac}       % compact symbols for 1/2, etc.
\usepackage{microtype}      % microtypography
\usepackage{lipsum}
\usepackage{fancyhdr}       % header
\usepackage{graphicx}       % graphics
\usepackage{enumitem}
\usepackage{amsmath}
\usepackage{tabularx}
\usepackage{multirow}
\graphicspath{{figs/}}     % organize your images and other figures under media/ folder

\newcommand{\blktitle}[1]{\noindent\textbf{#1\quad}}
\newcommand{\scoreone}{\protect\includegraphics[width=0.025\linewidth]{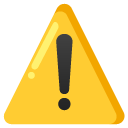}}
\newcommand{\scoretwo}{\protect\includegraphics[width=0.025\linewidth]{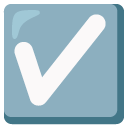}}
\newcommand{\scorethree}{\protect\includegraphics[width=0.025\linewidth]{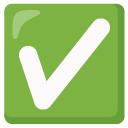}}

\hypersetup{
colorlinks=true,
linkcolor=black
}

\title{\protect\includegraphics[width=0.035\linewidth]{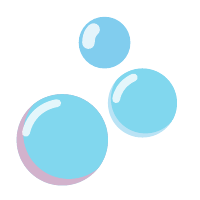} \texttt{OpenLens AI}: Fully Autonomous Research Agent for Health Infomatics}

\author{
Yuxiao Cheng\textsuperscript{1}~~~~~~~~ 
Jinli Suo\textsuperscript{1,2}\thanks{Corresponding author}\\
\textsuperscript{1}Department of Automation, Tsinghua University\\
\textsuperscript{2}Institute for Brain and Cognitive Science, Tsinghua University (THUIBCS)\\
\texttt{cyx22@mails.tsinghua.edu.cn, jlsuo@tsinghua.edu.cn} \\
}

\begin{document}
\maketitle

\begin{abstract}
Health informatics research is characterized by diverse data modalities, rapid knowledge expansion, and the need to integrate insights across biomedical science, data analytics, and clinical practice. These characteristics make it particularly well-suited for agent-based approaches that can automate knowledge exploration, manage complex workflows, and generate clinically meaningful outputs. Recent progress in large language model (LLM)-based agents has demonstrated promising capabilities in literature synthesis, data analysis, and even end-to-end research execution. However, existing systems remain limited for health informatics because they lack mechanisms to interpret medical visualizations and often overlook domain-specific quality requirements. To address these gaps, we introduce \texttt{OpenLens AI}, a fully automated framework tailored to health informatics. \texttt{OpenLens AI} integrates specialized agents for literature review, data analysis, code generation, and manuscript preparation, enhanced by vision-language feedback for medical visualization and quality control for reproducibility. The framework automates the entire research pipeline, producing publication-ready LaTeX manuscripts with transparent and traceable workflows, thereby offering a domain-adapted solution for advancing health informatics research.
\end{abstract}

% keywords can be removed
% \keywords{AI Agent \and Second keyword \and More}
\section{Introduction}
Health informatics research lies at the intersection of biomedical science, data analytics, and clinical practice. It is defined by heterogeneous clinical data, diverse methodological demands, and the need to translate findings into meaningful clinical insights. Researchers must handle data ranging from clinical time series to genomic information, while also navigating the rapid expansion of biomedical literature. The combination of high data complexity, multimodal integration, and fast-growing knowledge bases makes health informatics uniquely suited for approaches that can automate knowledge discovery and streamline research workflows.

Large language model (LLM)-based agents offer strong potential to address these challenges by automating labor-intensive tasks and enhancing analytical capacity\cite{wangScientificDiscoveryAge2023a, chenAI4ResearchSurveyArtificial2025, wangSurveyLLMbasedAgents2025}. Recent progress has rapidly expanded the landscape of AI research agents, demonstrating capabilities in idea generation \cite{garikaparthiIRISInteractiveResearch2025, siCanLLMsGenerate2024}, literature synthesis \cite{lalaPaperQARetrievalAugmentedGenerative2023, maSciAgentToolaugmentedLanguage2024}, and even end-to-end research execution \cite{konCurieRigorousAutomated2025, schmidgallAgentLaboratoryUsing2025}. These advances, underpinned by increasingly sophisticated reasoning abilities, suggest the feasibility of fully automated workflows that span hypothesis formulation to manuscript preparation\cite{konCurieRigorousAutomated2025, schmidgallAgentLaboratoryUsing2025, ifarganAutonomousLLMdrivenResearch2024, dengAutonomousSelfEvolvingResearch2025}, enabling faster and more scalable scientific discovery.

\begin{figure*}[t]
\centering
    \includegraphics[width=1.0\linewidth, trim=0.8in 3in 0.8in 1.5in, clip]{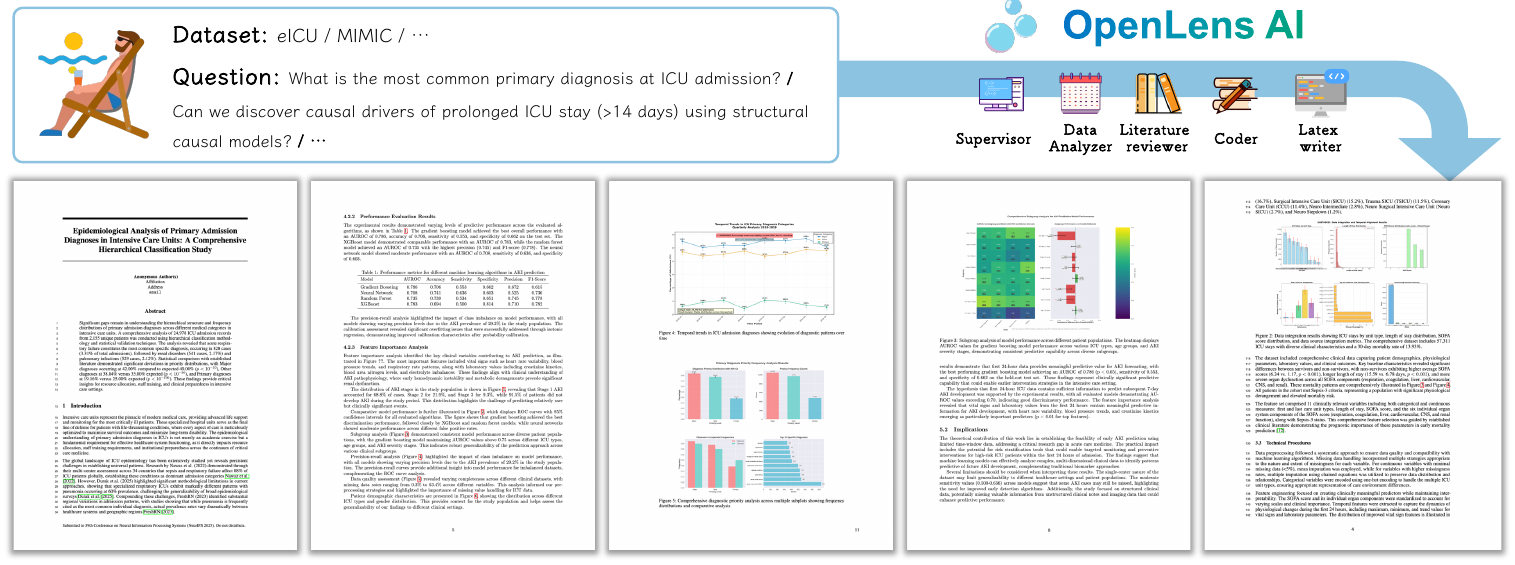}
    \caption{Example excerpts from manuscripts generated by OpenLens AI. Each excerpt illustrates structured sections, integrated figures, and coherent narrative flow. The system leverages vision-language feedback to ensure high-quality visual layouts and clear presentation of content.}
    \label{fig:example}
    \vspace{-5mm}
\end{figure*}

Despite this progress, existing research agents remain insufficient for health informatics. General-purpose systems often fail to capture two domain-specific needs. First, they lack vision-language feedback, which is crucial for interpreting complex medical visualizations—an essential component of health informatics analysis \cite{shiEHRAgentCodeEmpowers2024, shimgekarAgenticAIFramework2025}. Second, they seldom include systematic validation mechanisms tailored to medical contexts, raising the risk of plausible but misleading findings with serious implications for healthcare \cite{yuanKIDReviewKnowledgeGuidedScientific2022}.

To address these gaps, we introduce \texttt{OpenLens AI}, a fully automated framework purpose-built for health informatics research. \texttt{OpenLens AI} integrates domain-specific workflows, vision-language feedback, and quality control mechanisms to produce reliable, reproducible outputs with minimal human intervention. Unlike systems limited to markdown or plain-text reporting, it generates publication-ready LaTeX manuscripts with built-in checks to enhance transparency and credibility. Our key contributions are:

\begin{itemize}[leftmargin=*]
\item A modular agent architecture tailored to health informatics, featuring specialized agents for literature review, data analysis, code generation, and manuscript preparation.
\item Integration of vision-language feedback mechanisms to evaluate and refine visual outputs, including plots, charts, and curves—addressing a critical gap in current research agents.
\item A quality control framework that supports methodological soundness, statistical reliability, and reproducibility, reducing the likelihood of misleading results.
\item End-to-end automation of the health informatics research pipeline, from initial ideation to publication-ready LaTeX manuscripts.
\end{itemize}

\section{Related Works}

\blktitle{LLM agents for science.}
LLM agents have recently emerged as a prominent research topic, attracting attention for their potential to reshape how scientific discovery is conducted. They are increasingly explored as assistants that reduce the cognitive burden on researchers and accelerate the research cycle.  
They have been applied to idea mining, helping to spark novel hypotheses and uncover unexplored directions \cite{garikaparthiIRISInteractiveResearch2025, siCanLLMsGenerate2024, wangSciPIPLLMbasedScientific2025, wuSC4ANMIdentifyingOptimal2025}.  
In paper reading, agents condense contributions, extract methods, and surface limitations to support rapid understanding \cite{lalaPaperQARetrievalAugmentedGenerative2023, maSciAgentToolaugmentedLanguage2024, quRecursiveIntrospectionTeaching2024, singhalExpertlevelMedicalQuestion2025a}.  
For literature search, they identify and organize relevant works, enabling structured exploration of knowledge \cite{hePaSaLLMAgent2025, IntroducingDeepResearch, pinedoArZiGoRecommendationSystem2024}.  
Agents can also be used for reviewing, where they provide preliminary assessments of novelty, rigor, and coherence to assist peer evaluation \cite{darcyMARGMultiAgentReview2024, shinMindBlindSpots2025, yuanKIDReviewKnowledgeGuidedScientific2022}.  
Finally, in paper writing, they contribute to drafting, polishing, and ensuring stylistic consistency \cite{chengChartReaderUnifiedFramework2023, ifarganAutonomousLLMDrivenResearch2025}.  

Beyond these targeted applications, self-evolving and adaptive agents extend capabilities by continuously refining strategies and integrating new tools. They also enhance scientific communication, for example through automated poster generation and human–AI collaborative frameworks \cite{dengAutonomousSelfEvolvingResearch2025, jinSTELLASelfEvolvingLLM2025, pangPaper2PosterMultimodalPoster2025, gaoEmpoweringBiomedicalDiscovery2024}. Building on these advances, fully automated end-to-end research agents have been envisioned as the next step toward AI-augmented science.  

\blktitle{Fully-automated AI agents for research.}
Large language model (LLM)-based autonomous agents are increasingly capable of conducting end-to-end scientific investigations with minimal human intervention. They can generate hypotheses, design experiments, and analyze results while maintaining transparency and traceability \cite{ifarganAutonomousLLMdrivenResearch2024, dengAutonomousSelfEvolvingResearch2025, luAIScientistFully2024}. Some frameworks, such as Curie\cite{konCurieRigorousAutomated2025} or Agent Laboratory\cite{schmidgallAgentLaboratoryUsing2025}, enable fully-automated AI agnts for scientific research and integrate mechanisms to ensure experimental rigor and internal consistency. Virtual environments like DiscoveryWorld\cite{jansenDiscoveryWorldVirtualEnvironment2024} and AgentRxiv\cite{schmidgallAgentRxivCollaborativeAutonomous2025} allow continuously improvements upon prior research and systematic evaluation of agents’ discovery capabilities, revealing both their potential and the limitations imposed by simulation fidelity.

\blktitle{AI agents for medical research.}
In clinical contexts, AI agents are reshaping healthcare workflows by automating data interpretation, code generation, paper wrting, and advancing scientific discovery\cite{wangScientificDiscoveryAge2023a, wangSurveyLLMbasedAgents2025}. Recent work shows that agentic pipelines can not only conduct complex data analyses \cite{shiEHRAgentCodeEmpowers2024}, but also streamline entire clinical data pipelines \cite{shimgekarAgenticAIFramework2025}, and even self-evolve to acquire new capabilities in the course of research \cite{jinSTELLASelfEvolvingLLM2025}. Our \texttt{OpenLens AI} represents a fully automated agentic system specially tailored for medical research, requiring no human intervention throughout the research process. Unlike existing approaches, it generates well-formatted LaTeX papers with built-in academic rigor checks to minimize hallucinations, rather than relying on markdown or plain-text reports. On the other hand, it integrates vision-language feedback to enhance both visual quality and scientific rigor.  

\begin{figure*}[t]
\centering
    \includegraphics[width=1.0\linewidth, trim=1.4in 2in 1.4in 1in, clip]{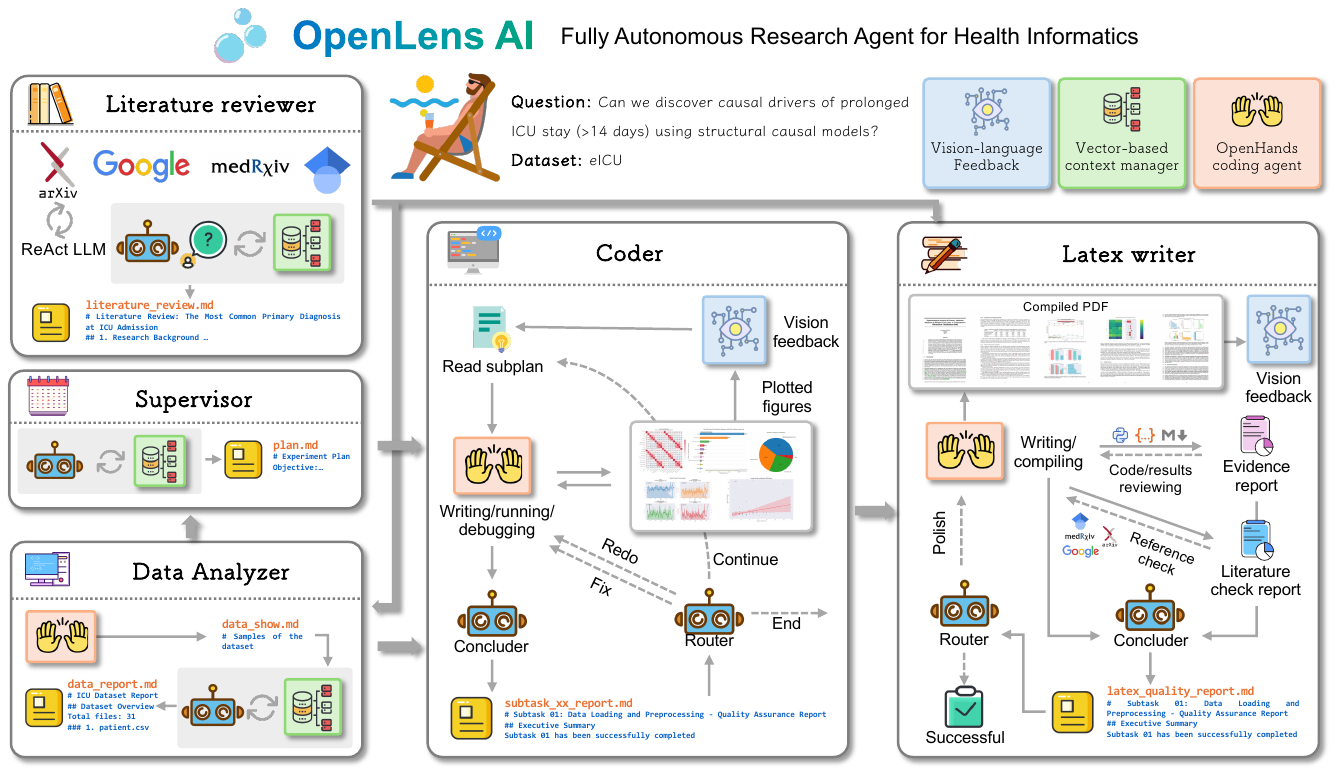}
    \caption{\textbf{Architecture of \texttt{OpenLens AI}.} The framework consists of five core modules: Supervisor (task planning and coordination), Literature Reviewer (literature search and synthesis), Data Analyzer (data processing and analysis), Coder (code generation and execution), and LaTeX Writer (manuscript preparation).}
    \label{fig:main_structure}
    \vspace{-5mm}
\end{figure*}

\section{Methods}

\texttt{OpenLens AI} is built upon a modular architecture in which autonomous agents are optimized for different stages of the scientific research pipeline. Instead of relying on a monolithic model, the framework adopts a distributed design where agents interact through a shared state representation under the supervision of a central controller. This modularity ensures scalability, fault tolerance, and flexibility, as each module can be independently extended or replaced without major modifications to the overall system.

The framework operates under the paradigm of ``research-as-process'' automation
\cite{saetraRiseResearchAutomaton2025}. A user provides a concise research idea, optionally accompanied by data, and the system decomposes this input into a sequence of subtasks. Each subtask is delegated to coder module, which executes it autonomously and records intermediate results. These subtasks collectively culminate in a comprehensive research report, including literature review, data analysis, methodology, and synthesized findings. To ensure robustness, the framework employs LangGraph\cite{LangGraph} as workflow engine that enforces a directed graph-based execution structure, manages retries in case of failures, and maintains a persistent record of intermediate states. Moreover, our coder module is built upon OpenHands\cite{wangOpenHandsOpenPlatform2024}, which provides low-level state management and seamless integration with coding tools.

\subsection{Specialized Modules}

\blktitle{Supervisor.}
The supervisor serves as the global coordinator of the research workflow. Its primary function is to decompose a user query into structured subtasks, specifying for each subtask the expected inputs and outputs. By clearly defining the task boundaries and data requirements, the Supervisor enables downstream agents to execute their responsibilities in a well-organized and consistent manner. This structured decomposition ensures that each subtask can operate autonomously while adhering to the overall research plan, facilitating traceability and interpretability throughout the pipeline.

\blktitle{Literature reviewer.}
The literature reviewer constructs an autonomous pipeline for surveying and synthesizing relevant literature, leveraging a ReAct-based reasoning framework \cite{yaoReActSynergizingReasoning2023}. Given a research query, the system first initializes a search-oriented language model $\mathcal{M}_\text{search}$ and a report-writing language model $\mathcal{M}_\text{write}$, then activates a suite of domain-specific search tools $\mathcal{T} = \{\text{ArXiv, MedRxiv, Tavily},\dots\}$ to retrieve candidate papers. Each retrieval step feeds into the ReAct agent, which iteratively decides between \textit{reasoning} and \textit{action} by invoking appropriate tools until a stopping criterion based on the number of tool calls is met. This can be summarized as
\[
\text{state}_{t+1} = f_\text{ReAct}(\text{state}_t, \mathcal{T}, \mathcal{M}_\text{search}), \quad \text{stop if } \#\text{tool calls} \ge k_\text{min}.
\]

Upon reaching the limit, the pipeline transitions $\rightarrow$ the report-writing chatbot, which consolidates retrieved knowledge into a coherent document $\mathcal{R}$, merging redundancies and emphasizing novel contributions. The resulting report is then formalized through a report-writing tool and finally stored in a persistent state, enabling downstream consumption. The overall workflow can be visualized as:  
\[
\text{START} \rightarrow \text{literature search} \xrightarrow{\text{tool calls}} \text{write report} \rightarrow \text{report tool node} \rightarrow \text{END}.
\]

This design ensures the literature review is dynamically adaptive, technically traceable, and both comprehensive and concise, bridging multi-source knowledge into a structured, actionable summary.

\blktitle{Data analyzer.}
The data analyzer orchestrates a multi-stage pipeline that converts raw datasets into structured, interpretable insights. Initially, a primary language model $\mathcal{M}_\text{base}$ is initialized to interface with various analysis tools, while a router model $\mathcal{M}_\text{router}$ supervises decision-making and adaptive workflow branching. Raw inputs, ranging from clinical time-series to tabular records, are first processed by a coding tool $\mathcal{C}$, which generates executable analysis scripts and performs validation through execution checks. The results $\mathcal{D}_\text{processed}$ are then incorporated into a chatbot-driven narrative agent that produces descriptive reports $\mathcal{R}$, integrating visualizations, statistical summaries, and natural language interpretations.

Workflow control is governed by conditional routing: the router agent monitors outputs and keywords to determine whether to return for reprocessing, or proceed to final report generation. Formally, the state update with a stopping condition can be expressed as
\[
\text{state}_{t+1} =
\begin{cases}
f_\text{OpenHands}(\text{state}_t), & \text{if data missing or } R(\text{state}_t) = \text{\textit{return}},\\
f_\text{LLM analysis}(\text{state}_t), & \text{if data ready and } R(\text{state}_t) = \text{\textit{continue}}.
\end{cases}
\]

The overall pipeline can be visualized more clearly in two stages:
\[
\text{START} \rightarrow \text{OpenHands node} \xrightarrow{\text{data processed}} \text{LLM analysis/Redo} 
\]
\[
\rightarrow \text{Report generation} \xrightarrow{\text{router decision}} \text{END/loop back}.
\]

This design enables iterative, adaptive data analysis where preprocessing, coding, visualization, and narrative synthesis are tightly coupled. By combining automated execution with natural language reasoning, the system produces reproducible and interpretable insights while maintaining flexibility for complex datasets.

\blktitle{Coder.}
The coder transforms high-level experimental plans into executable code through an iterative pipeline that tightly couples code generation, validation, and adaptive refinement. Initially, a plan-reading agent extracts subtasks from the overall experimental specification, initializing the subtask index $i$ and guiding subsequent execution. For each subtask, a coding agent $\mathcal{C}_\text{openhands}$ generates executable scripts, incorporating feedback from prior failures if present. Generated code is then validated, not only through automated execution checks but also via a vision-language model $\mathcal{V}$ that evaluates any visual outputs such as plots, providing feedback to improve clarity and correctness.

Workflow control is orchestrated by a router agent $\mathcal{R}$, which monitors tool outputs, subtask completion, and vision feedback. The state evolution can be formally expressed as
\[
\text{state}_{t+1} =
\begin{cases}
f_\text{continue}(\text{state}_t), & \text{if } R(\text{state}_t) = \text{\textit{continue next subtask}},\\
f_\text{redo}(\text{state}_t), & \text{if } R(\text{state}_t) = \text{\textit{redo last subtask} \textbf{or} \textit{validation fails}},\\
f_\text{fix}(\text{state}_t), & \text{if } R(\text{state}_t) = \text{\textit{fix last subtask}},\\
\text{STOP}, & \text{if } R(\text{state}_t) = \text{\textit{alter plan} \textbf{or} \textit{all subtasks completed}}.
\end{cases}
\]

The overall pipeline can be visualized in two stages to clarify iteration and routing:
\[
\text{START} \rightarrow \text{Plan reader} \rightarrow \text{OpenHands coding} \rightarrow \text{Vision-language validation}
\]
\[
\rightarrow \text{Concluder} \xrightarrow{\text{router decision}} \text{Subtask Continue / restart / fix} \rightarrow \text{END}.
\]

This architecture enables robust, adaptive code generation, ensuring that computational procedures are correct, scientifically meaningful, and refined based on both execution and visual feedback, while supporting iterative subtask management and dynamic workflow branching.

\blktitle{LaTeX writer.}
The LaTeX writer orchestrates the assembly of a complete scientific manuscript by integrating outputs from all preceding agents. It initializes with a state-clearing node to reset prior messages and counters, then collects available figures and datasets $\mathcal{F}$, ensuring only VLM-approved images are included. Each manuscript section: Introduction, Related Works, Methods, and Experiments is iteratively drafted by a coding agent $\mathcal{C}_\text{openhands}$, which incorporates previous feedback to refine textual and visual content. A validator agent then evaluates the generated LaTeX code and compiled PDF, leveraging a vision-language model $\mathcal{V}$ to identify elements requiring improvement.

Workflow is controlled by a router agent $\mathcal{R}$, which interprets tool outputs, polish counters, and VLM feedback to determine the next action. The state update can be formalized as
\[
\text{state}_{t+1} =
\begin{cases}
f_\text{polish}(\text{state}_t), & \text{if } R(\text{state}_t) = \text{\textit{polish} \textbf{and} \textit{polish limit not reached}},\\
\text{STOP}, & \text{if } R(\text{state}_t) = \text{\textit{end} \textbf{or} \textit{polish limit reached}}.
\end{cases}
\]

The overall pipeline can be visualized in two stages to clarify the sequential and iterative process:
\[
\text{START} \rightarrow \text{Collect figures} \rightarrow \text{Write sections (Intro, related, ...)}
\]
\[
\rightarrow \text{Validator} \xrightarrow{\text{router decision}} \text{Concluder chatbot / tools} \rightarrow \text{END / Loop for polish}.
\]

This architecture ensures that the final manuscript maintains academic rigor, integrates visual and textual feedback, and allows for iterative refinement until publication-quality output is achieved.

Although each module, i.e., literature reviewer, data analyzer, coder, and LaTeX writer, operates autonomously, effective automation relies on a shared state that tracks the research plan, subtask progress, and intermediate outputs, including textual and visual artifacts. Agents read from and write to this state iteratively, enabling reproducibility, interpretability, and incremental refinement. Workflow coordination is enforced by router agents, which conditionally direct the flow based on tool outputs, validation results, and vision-language feedback. Unsatisfactory outputs trigger automatic retries, subtask reprocessing, or targeted improvements, ensuring that errors are contained and corrected without disrupting the overall pipeline. This architecture guarantees robustness, supports iterative refinement, and maintains consistent progress toward a publication-quality manuscript.

\subsection{Quality Control}

\blktitle{Academic rigor check.}
The framework incorporates systematic procedures to verify the methodological soundness of each experiment. It automatically checks for common pitfalls such as data leakage, unrealistic performance metrics, and improper handling of temporal information in time-series data. It also audits feature engineering pipelines to ensure that only information available at prediction time is used. These safeguards reduce the risk of spurious or irreproducible findings.  

\blktitle{Evidence traceability check.}
Every claim or result in the manuscript is linked to its underlying evidence, including datasets, scripts, and experiment logs. The system generates a structured traceability report mapping manuscript paragraphs to their supporting files and code segments. This ensures transparency, enables reproducibility, and provides a clear audit trail for validation of the research process.  

\blktitle{Literature check.}
The framework integrates an academic literature verification module that validates all cited references. It cross-checks each reference against authoritative sources to confirm correctness of metadata (authors, titles, venues, years, DOIs), removes fabricated or unverifiable entries, and enforces consistency in formatting. This process ensures both the accuracy of citations and the reliability of the scholarly context in which results are situated.  

\blktitle{Vision-language feedback.}
To ensure the quality of outputs, the framework integrates vision-language models at critical stages, particularly in data visualization and manuscript preparation. These models provide perceptual feedback on generated artifacts that are difficult to evaluate in text alone, such as plots or figures. Combined with iterative self-refinement loops, this multimodal evaluation pipeline enhances both the readability and scientific validity of results. By embedding quality assurance into every stage of the workflow, the framework reduces error accumulation and improves the overall rigor of automated research.

\section{Experiments}

Since the proposed task is novel, there are no directly comparable baseline methods. To evaluate the effectiveness of \texttt{OpenLens AI}, we designed a benchmark consisting of 18 tasks with increasing levels of difficulty, covering the full pipeline from literature review to manuscript generation. Considering that medical research often requires private deployment due to confidentiality concerns, we employed medium-scale models to balance performance and deployability: GLM-4.5-Air \cite{teamGLM45AgenticReasoning2025} as the language model backbone, and GLM-4.1V-9B-Thinking \cite{hongGLM41VThinkingVersatileMultimodal2025} as the vision-language model for visual reasoning. For efficient benchmarking, we restrict the maximum number of subtask refinements and LaTeX polishing iterations to two each. All code, benchmarks, and evaluation methods have been open-sourced and are available at \url{https://github.com/jarrycyx/openlens-ai}.

\blktitle{Benchmark.}
We selected two widely used, openly available clinical datasets to test our system: 
MIMIC-IV \cite{johnsonMIMICIVFreelyAccessible2023a} and eICU \cite{johnsonEICUCollaborativeResearch, pollardEICUCollaborativeResearch2018a}. 
These resources were chosen because they represent complementary large-scale critical care databases: MIMIC-IV provides rich ICU patient records from a single hospital system, while eICU offers a multi-center perspective covering diverse hospital settings. 
To ensure feasibility of repeated end-to-end evaluation within reasonable runtime, we used only the ICU subset of MIMIC-IV and the demo version of eICU (approximately 2,500 unit stays). 
This setup provides sufficient complexity and heterogeneity to challenge the agents, while remaining computationally tractable for iterative testing.

\blktitle{Evaluation.}
The outputs of the system were evaluated using an LLM-as-Judge protocol, which provides neutral, reproducible, and fine-grained assessment. 
Each task was analyzed along five dimensions: (1) plan completion, (2) code execution, (3) result validity, (4) paper completeness, and (5) conclusion quality. The LLM-as-Judge are open-sourced as well.
Each dimension is scored on a three-point scale, with scores represented by emoji symbols in our tables: 
\scoreone (1 = severe issues that make the research fundamentally wrong), 
\scoretwo (2 = moderate issues that still allow the research to be valid), and 
\scorethree (3 = minor or no issues). 
This scoring design captures both catastrophic failures and subtle imperfections. 
The benchmark covered easy (E1–E3), medium (M1–M3), and hard (H1–H3) tasks, with evaluation results summarized in Table~\ref{tab:evaluation} and task definitions listed in Table~\ref{tab:questions}.

\blktitle{Results.}
Examples of the generated manuscripts are shown in Fig. \ref{fig:example}. Thanks to the integrated vision-language feedback, the LaTeX writer produces manuscripts with visually appealing formatting, properly sized and positioned figures, and consistent section structures. This ensures that even complex outputs maintain clarity and readability, enhancing both the interpretability and professional presentation of the generated papers.
Overall, the evaluation demonstrates that \texttt{OpenLens AI} is able to complete the majority of tasks with high reliability, especially for low- to medium-difficulty questions. 
For easy tasks (E1–E3), the system consistently achieved high scores across all five dimensions, with most results reaching \scorethree, indicating that the pipeline can reliably handle descriptive statistics and straightforward clinical queries. 
For medium tasks (M1–M3), performance remained strong but exhibited variability: issues were observed mainly in code execution and result validity, where occasional errors in data preprocessing or model fitting led to lower scores (\scoretwo). 
This shows that while the agent can still deliver valid research outputs, intermediate steps may require refinement. 
For hard tasks (H1–H3), the system struggled more substantially. Severe issues (\scoreone) appeared in causal discovery and generalization experiments, reflecting the intrinsic challenges of such tasks and the limitations of automated reasoning without additional domain priors. 

These results suggest a clear performance gradient: the framework is robust for routine statistical analysis and exploratory studies, moderately reliable for predictive modeling under controlled conditions, and still limited when confronted with open-ended causal or generalization questions. 
Nevertheless, even in hard cases, the generated manuscripts remain structurally coherent, with the LaTeX writer consistently producing publication-style outputs.

\begin{table}[t]
\centering
\caption{Benchmark task list}
\label{tab:questions}
\small
\begin{tabular}{cp{12cm}}
\toprule
ID & Problem Description \\
\midrule
E1 & What is the distribution of patient ages, and how does it differ between male and female patients? \\
E2 & What is the in-hospital mortality rate for patients admitted with pneumonia? \\
E3 & What is the most common primary diagnosis at ICU admission? \\
\midrule
M1 & How do missingness patterns in lab tests influence the bias of sepsis prediction models? \\
M2 & How accurately can 30-day mortality be predicted using vital signs and lab values recorded in the first 24 hours of ICU stay? \\
M3 & What is the effect of age and comorbidity count on sepsis mortality? \\
\midrule
H1 & Can we discover causal drivers of prolonged ICU stay (>14 days) using structural causal models? \\
H2 & How do hospital-level differences (staff ratio, region) confound predictive modeling of mortality in eICU? \\
H3 & How do predictive models trained generalize to older patients (>75 years) compared to younger ones? \\
\bottomrule
\end{tabular}
\end{table}

\begin{table}[t]
\centering
\small
\caption{Evaluation Results with GLM-4.5 \cite{teamGLM45AgenticReasoning2025}. Each criterion is rated on a 3-point scale: \scoreone (1 = severe issues that make the research fundamentally wrong), \scoretwo (2 = moderate issues that still allow the research to be valid), \scorethree (3 = minor or no issues).}
\label{tab:evaluation}
\begin{tabular}{lllcccccc}
\toprule
Difficulty & ID & Dataset & Plan & Code & Result & Paper & Conclusion & Avg \\
\midrule
\multirow{6}{*}{Easy} & \multirow{2}{*}{E1} & eICU (Demo)    & \scorethree & \scorethree & \scorethree & \scorethree & \scorethree & 3.0 \\
                      &                     & MIMIC IV (ICU) & \scoretwo & \scoretwo & \scoretwo & \scorethree & \scorethree & 2.4 \\
\cmidrule{2-9}
                        & \multirow{2}{*}{E2} & eICU (Demo)    & \scorethree & \scorethree & \scorethree & \scoretwo & \scorethree & 2.8 \\
                      &                     & MIMIC IV (ICU) & \scorethree & \scorethree & \scorethree & \scorethree & \scorethree & 3.0 \\
\cmidrule{2-9}
                        & \multirow{2}{*}{E3} & eICU (Demo)    & \scoretwo & \scorethree & \scorethree & \scoretwo & \scorethree & 2.6 \\
                      &                     & MIMIC IV (ICU) & \scorethree & \scorethree & \scorethree & \scoretwo & \scorethree & 2.8 \\
\midrule
\multirow{6}{*}{Medium} & \multirow{2}{*}{M1} & eICU (Demo)    & \scorethree & \scoretwo & \scorethree & \scoretwo & \scorethree & 2.6 \\
                        &                     & MIMIC IV (ICU) & \scorethree & \scorethree & \scorethree & \scorethree & \scorethree & 3.0 \\
\cmidrule{2-9}
                        & \multirow{2}{*}{M2} & eICU (Demo)    & \scoretwo & \scorethree & \scoreone & \scoretwo & \scoreone & 1.8 \\
                        &                     & MIMIC IV (ICU) & \scorethree & \scorethree & \scoretwo & \scoretwo & \scorethree & 2.6 \\
\cmidrule{2-9}
                        & \multirow{2}{*}{M3} & eICU (Demo)    & \scoretwo & \scorethree & \scoretwo & \scorethree & \scoretwo & 2.4 \\
                        &                     & MIMIC IV (ICU) & \scoretwo & \scorethree & \scorethree & \scoretwo & \scorethree & 2.6 \\
\midrule
\multirow{6}{*}{Hard} & \multirow{2}{*}{H1} & eICU (Demo)    & \scoretwo & \scorethree & \scoretwo & \scorethree & \scoretwo & 2.4 \\
                      &                     & MIMIC IV (ICU) & \scoretwo & \scoretwo & \scoreone & \scoretwo & \scoreone & 1.6 \\
\cmidrule{2-9}
                        & \multirow{2}{*}{H2} & eICU (Demo)    & \scorethree & \scorethree & \scorethree & \scorethree & \scorethree & 3.0 \\
                      &                     & MIMIC IV (ICU) & \scoreone & \scoretwo & \scoreone & \scorethree & \scoretwo & 1.8 \\
\cmidrule{2-9}
                        & \multirow{2}{*}{H3} & eICU (Demo)    & \scorethree & \scorethree & \scoretwo & \scorethree & \scorethree & 2.8 \\
                      &                     & MIMIC IV (ICU) & \scorethree & \scorethree & \scoretwo & \scorethree & \scoretwo & 2.6 \\
\bottomrule
\end{tabular}
\end{table}

\section{Conclusion.}
In this work, we present \texttt{OpenLens AI}, an autonomous multi-agent system designed to perform end-to-end medical research tasks, from literature review to manuscript generation. By integrating specialized modules, including a supervisor, literature reviewer, data analyzer, coder, and LaTeX writer, \texttt{OpenLens AI} achieves coherent coordination through a shared state and conditional workflow routing. Our experiments on a novel benchmark composed of 9 tasks spanning easy, medium, and hard difficulty levels demonstrate that the system can produce scientifically meaningful outputs, generate valid code, and compile publication-quality manuscripts. The inclusion of vision-language feedback further improves visual clarity and manuscript formatting, ensuring that figures and tables are both accurate and aesthetically consistent. Evaluation using an LLM-as-Judge protocol shows that the system maintains high scores across multiple dimensions, including plan completion, result validity, and conclusion quality, highlighting its robustness and adaptability in complex medical research scenarios.

\blktitle{Limitations and Future Work.}
Despite promising results, the current study has several limitations. First, we have not conducted rigorous comparative experiments against general-purpose agent systems, due to the lack of directly comparable benchmarks and differences in task scope. Second, the system relies on medium-scale models to balance deployability and performance, which may constrain its capabilities when handling highly complex or multimodal datasets. Third, our current evaluation focuses primarily on selected subsets of MIMIC-IV and eICU datasets; broader validation across diverse clinical data is necessary to fully assess generalizability.

Future work will address these limitations by constructing more comprehensive and publicly accessible benchmarks tailored to medical research tasks, enabling standardized performance comparisons across different agent architectures. We also plan to explore fine-tuning dedicated models for both language and vision-language components, optimizing them for accuracy, efficiency, and privacy preservation. Such efforts aim to facilitate deployment in medical institutions with strict data confidentiality requirements, ensuring that autonomous research agents can be safely and effectively used in real-world clinical research settings. Collectively, these directions will advance the development of reliable, privacy-conscious, and high-performing autonomous agents for medical science.

\section*{Acknowledgments}
This work is jointly supported by the National Key R\&D Program of China (Grant No. 2024YFF0505703) and the National Natural Science Foundation of China (Grant No.  62088102).

%Bibliography
\bibliographystyle{unsrt}  
\bibliography{ref}

\end{document}